\documentclass[conference]{IEEEtran}
\usepackage[T1]{fontenc}
\IEEEoverridecommandlockouts
\usepackage{cite}
\usepackage{amsmath,amssymb,amsfonts}
\usepackage{algorithmic}
\usepackage{graphicx}
\usepackage{textcomp}
\usepackage{xcolor}
\usepackage{algorithm}
\def\BibTeX{{\rm B\kern-.05em{\sc i\kern-.025em b}\kern-.08em
    T\kern-.1667em\lower.7ex\hbox{E}\kern-.125emX}}
\begin{document}

\title{CogniPlay: a work-in-progress Human-like model for General Game Playing
\thanks{GameTable COST CA22145 Action}
}

%
\author{\IEEEauthorblockN{Rautureau Aloïs}
\IEEEauthorblockA{
\textit{École Normale Supérieure de Rennes}\\
Rennes, France \\
0009-0007-8507-2167}
\and
\IEEEauthorblockN{Piette Éric}
\IEEEauthorblockA{\textit{ICTEAM} \\
\textit{Université Catholique de Louvain}\\
Louvain-La-Neuve, Belgium \\
0000-0001-8355-636X}
}

\maketitle

\begin{abstract}
While AI systems have equaled or surpassed human performance in a wide variety of games such as Chess, Go, or Dota 2, describing these systems as truly "human-like" remains far-fetched. Despite their success, they fail to replicate the pattern-based, intuitive decision-making processes observed in human cognition. This paper presents an overview of findings from cognitive psychology and previous efforts to model human-like behavior in artificial agents, discusses their applicability to General Game Playing (GGP) and introduces our work-in-progress model based on these observations: CogniPlay.
\end{abstract}


\section{Introduction}
Although AI systems have surpassed human performance in games such as Chess \cite{Campbell2002}, Go \cite{Silver2016}, and competitive games like Dota 2 \cite{openai2019dota2largescale}, describing them as "human-like" would be an overstatement. Despite their exceptional performance, these systems fail to accurately replicate the selective, pattern-based decision-making that characterizes human cognition \cite{DeGroot1978, Chase1973}. 

Designing agents that model or mimic this behavior could have a wide range of applications. In online environments such as video games, human-like agents can replace disconnected players in multiplayer settings or enhance the realism of non-player characters (NPCs) \cite{Milani2023}. The perceived human-likeness of an agent has also been recognized as a crucial factor in cooperative environments \cite{Carroll2019}, driving interest in such models within the field of Human-Robot Interaction (HRI) \cite{Turnwald2019}. Beyond interactive applications, human-like agents can serve as experience gatherers, accumulating strategic insights that can later be distilled for human use. Unlike super-human agents, which often rely on strategies that disregard cognitive and memory limitations inherent to human cognition, human-like agents would offer replicable approaches, making their strategies more accessible for practical application.

This work explores existing methods for developing human-like agents and examines their applicability to General Game Playing (GGP), a field that, to our knowledge, has not yet been studied from this perspective. GGP aims to create agents capable of playing games effectively using only a formal game description, with limited time for familiarization with the environment \cite{Genesereth2005}. 
Adapting such methods for GGP would offer greater control over an agent’s strength and playstyle addressing a common issue with strong AI opponents—namely, the artificial tuning of difficulty through methods such as depth limiting or reduced search time. Such models would also enable the automatic generation of more believable game-playing data, benefiting tabletop game platforms and designers by automating playtesting for early prototypes.

This last point would also be particularly valuable for reconstructing traditional games within the Ludii game-playing system \cite{Piette2020}. It currently relies on Monte Carlo Tree Search (MCTS) \cite{Kocsis2006} and expert iteration \cite{Anthony2017} to produce plausible gameplay data for partially recovered tabletop games \cite{Browne2019, Browne2020}. While this approach has shown promise \cite{cristComputationalApproachesRecognising2023}, its plausibility could be enhanced by incorporating a model that considers psychological and cognitive factors, as well as diverse motivations beyond merely winning, to more accurately replicate human gameplay.

In Section \ref{highlevel}, we expose key observations from psychology and cognitive science that outline features of human decision-making, as well as previous works applying these observations to various fields. This helps identify the gap between state-of-the-art game-playing agents and human players, which our work aims to fill, and motivate our model through past results. Section \ref{model} introduces CogniPlay, our work-in-progress model for GGP, based on the Double-Process Theory of Cognition (DPTC), a chunk-based representation of the game state and a selective search algorithm.

\section{Features of human decision-making} \label{highlevel}

Unlike machines, which process millions of board states per second and store large game trees in memory, humans are inherently limited in both memorization and processing speed. 

Our memorization of game states is optimized through a \textit{chunking} mechanism \cite{degrootHetDenkenVan1946, Chase1973, Reitman1976, Wolff1984, Gobet1998, Gobet2000}, grouping related pieces—connected by attack, defense, spatial proximity, or similarity in color or type—as a single perceptual unit.

Levinson and Snyder successfully implemented this approach in the MORPH Chess engine \cite{Levinson1991}. They represent board states as graphs of attack-defend relationships between pieces, deriving patterns from graph cuts associated to learned estimations of their Minimax evaluation. Despite being weaker than GnuChess (rated 1600 Elo), MORPH's learning rate and playstyle were described as promising.

A major challenge of their approach was combining multiple pattern weights. Models able to learn non-linear functions, such as Deep Neural Networks (DNNs) have been applied to Connect-4 \cite{Mandziuk2011}, incorporating spatial-locality and a pattern-based representation. Atomic patterns derived from game rules are grouped into overlapping spatial frames, each processed by a shared-weight neural network. The outputs are then combined in a final hidden layer, generating a policy that significantly outperformed classical architectures in predicting optimal moves without search.

As a result of our low memorization capabilities, we were also found to perform shallow searches, plateauing an average depth of 4.8 to 5.3 plies for expert Chess players ($\simeq$2000 Elo) \cite{degrootHetDenkenVan1946, charnessSearchChessAge1981, gobetPatternrecognitionTheorySearch1997a}. These searches are also highly selective, as human players prune a large chunk of the game tree intuitively instead of exploring it exhaustively \cite{degrootHetDenkenVan1946}.

This aligns with Kahneman's Double-Process Theory of Cognition (DPTC) \cite{Kahneman2003}. This theory posits that humans employ two different and co-dependent processes of thinking:
\begin{itemize}
    \item \textbf{System 1 (intuitive brain)} is fast, biased, based on intuition and the recognition of memorized patterns.
    \item \textbf{System 2 (analytical brain)} is slow, based on logical inference.
\end{itemize}

The interaction between these systems explains key aspects of human decision-making. In familiar situations—like responding to known Chess openings—, System 1 provides immediate responses. For novel situations, System 1 triggers System 2 for deeper analysis. This analysis is then gradually internalized by the intuitive brain. Importantly, once internalized, patterns become difficult to override, a mechanism that plays a key role in explaining cognitive biases in human decision-making.

Efforts to adapt tree search algorithms like Minimax \cite{Shannon1950} and MCTS \cite{Kocsis2006} to better reflect this aspect of human cognition typically incorporate a policy heuristic—often a neural network—to direct a tree search algorithm. We refer to these approaches as \textit{focused search} algorithms.

One such adaptation is focused Minimax search \cite{moriarty:focus}, where a neural network is used to discard a subset of actions at each depth of the search. In experiments using Othello, focused Minimax was found to outperform full-width search, but no comments were made on the resulting agent's playing style.

A similar idea was applied by He et al. \cite{He2024} to improve proactivity in Large Language Model (LLM) conversational agents. Their approach, explicitly based on DPTC, combines a small language model for policy evaluation with an MCTS-based search. Experimental results, including both quantitative metrics and human feedback, showed their system outperforming a competitive policy planner for conversational agents, Plug-and-Play Dialogue Policy Planner (PPDPP) \cite{deng2024plugandplay}, across three proactive conversation datasets.

Focused search and pattern-based approaches can be extended to GGP agents using Spatial State-Action Features \cite{Soemers2022a}. This approach has the advantage of generality—iteratively deriving features from base game rules—and interpretability, as learned features remain extractible \cite{Soemers2023a}, providing a generalized chunked game representation for GGP agents to direct an already general MCTS algorithm.

An important aspect of adversarial and cooperative environments, such as two-player zero-sum games or negotiations, is that the decision-making processes of both players are not necessarily symmetrical. The key psychological concept relevant to this asymmetry is Theory of Mind (ToM)—the ability to infer and model another person's beliefs, intentions and thought processes \cite{premack1978does}. In the context of games, ToM allows players to anticipate strategies or exploit perceived weaknesses in an opponent, referred to as adversarial problem solving.

Thagard provides an overview of adversarial problem solving \cite{thagard1992adversarial}, identifying key principles that emerge in domains such as military strategy, competitive games, and business negotiations. Central to these principles is the construction and continuous refinement of an opponent model—a representation of the opponent’s goals, knowledge, and past actions.

Jansen \cite{jansenUsingKnowledgeOpponent1992a} explored opponent-model search through a modified Minimax procedure that incorporates meta-knowledge of an opponent's search algorithm, derived from statistical observations. This approach was later extended into a probabilistic framework \cite{Donkers2001}, using predefined opponent models and inferring which mixture of these models best predicts the opponent's actions.

However, these methods all rely on extensive domain knowledge to produce opponent models, rarely available to GGP agents. To our knowledge, no effort has been made to generalize opponent modeling. We identify it as a promising research direction for human-like agents, but its practical implementation in GGP systems requires substantial additional research.

\section{A preliminary "human-like" model for GGP} \label{model}
\begin{figure}
    \centering
    \includegraphics[width=0.9\linewidth]{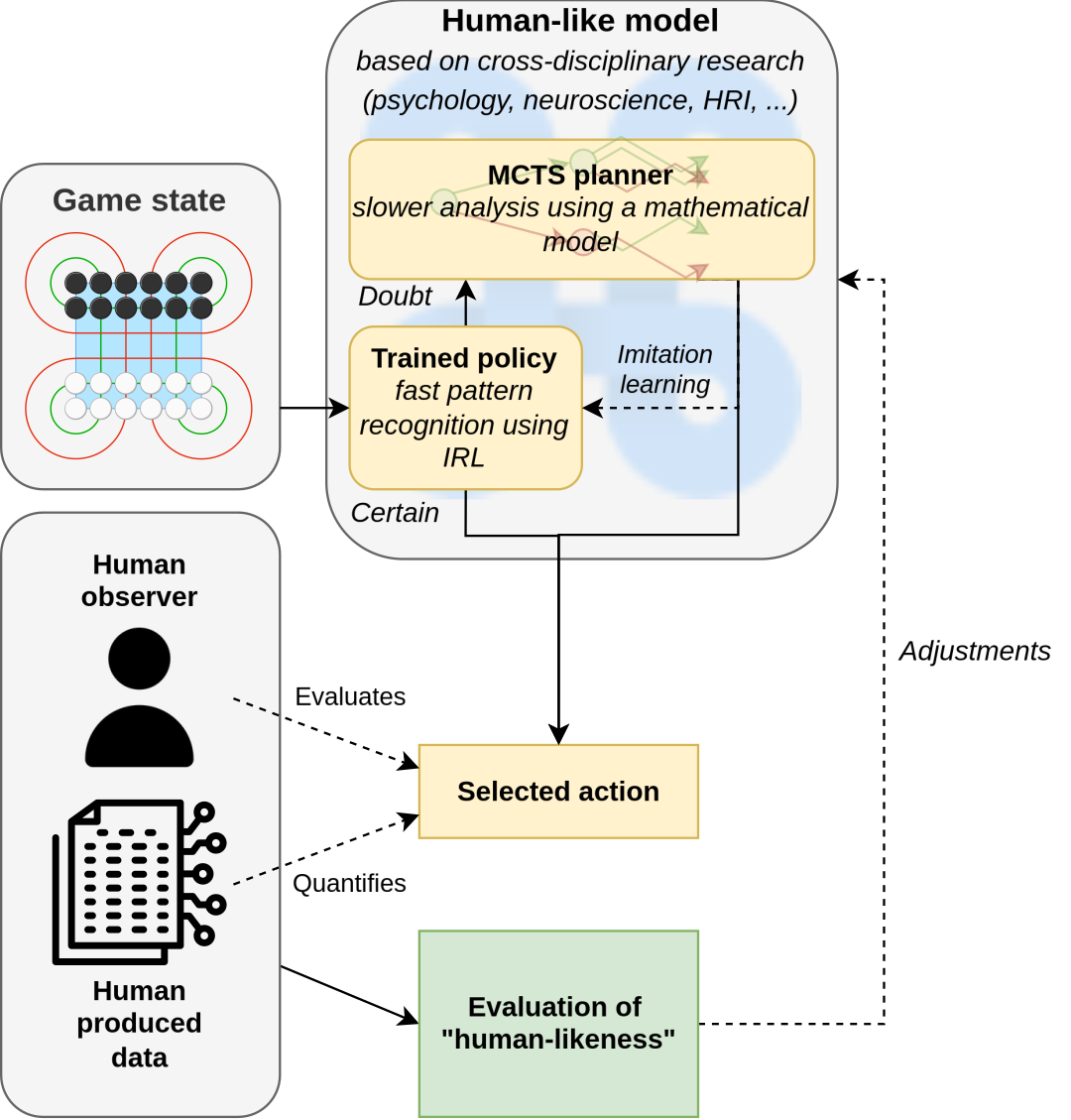}
    \caption{Our model and experimental setup for a human-like agent in GGP. \textit{Top}: The model is inspired by the Double Process Theory of Cognition, with System 1 relying on State-Action Features for action-set partitioning, and defaulting to System 2 (focused memory-bounded Monte-Carlo Tree Search) when encountering states of uncertainty. System 1 can learn from System 2 automatically through an expert iteration process. \textit{Bottom}: We assess the human-likeness of the model both qualitatively using Turing tests and quantitatively using move matching to ensure a robust measure of its performance.}
    \label{fig:model}
\end{figure}

Based on these observations and previous works, we propose a preliminary model for human-like GGP agents inspired by DPTC: CogniPlay (Figure \ref{fig:model}). It combines both pattern-driven, chunk-based methods for \textit{intuitive decisions} (System 1) and selective search for \textit{analytical decisions} (System 2), in the form of action-set partitioning using Spatial Action-State Features (Section \ref{intuitive}), and selective memory-bounded MCTS (Section \ref{selective_mcts}) respectively. 
Training is done using an expert iteration framework \cite{Anthony2017} where System 1 learns from the policy deduced by System 2 in novel situations.

\subsection{Intuitive decisions} \label{intuitive}
The intuitive brain (System 1) uses the policy derived from Spatial Action-State Features to drive the analytical brain (System 2) in its search. This is done through the process of action-set partitioning. System 1's heuristic is used to partition the set of legal actions into two subsets: intuitively good actions and intuitively bad actions. Only the set of intuitively good actions is searched further using System 2, aggressively pruning the search tree in an effort to mimic the selective search employed by human players. If only one action is deemed good by the intuitive brain, the analytical brain is bypassed entirely. We then deem the policy \textit{certain}, as opposed to \textit{doubtful} in the former case.

System 1 is trained using expert iteration only on states where it exhibits a doubtful policy. In such situations, the policy inferred by System 2 is used as a training sample for Spatial Action-State Features. While suboptimal, we expect this method to allow the emergence of \textit{biases} in the model.

This method allows training of the model without the need for human-generated data. It aims to approach the high-level human-like features of decision-making algorithmically rather than emulating them through imitation learning, serving a strong argument for the generalizability of the model across a wide variety of games, even when human-generated data is not sufficiently available.

\subsection{Selective Memory-Bounded Monte-Carlo Tree Search} \label{selective_mcts}
We aim to replicate the limits of human cognition in games, notably our limited memory and computational power, through the implementation of a memory-bounded MCTS algorithm \cite{powleyMemoryBoundedMonte2017a} guided by System 1's selective process.

This method slightly weakens the playing strength of MCTS agents, although no measurements have yet been made on its impact to their playing style. This scheme is better aligned with human-like planning, incorporating memory constraints and progressive deepening through a best-first iterative search procedure \cite{degrootHetDenkenVan1946}. It also introduces an additional parameter—memory bounds—that can be adjusted to influence both playing strength and style.

\subsection{Model evaluation}
Accurately measuring the human-likeness of a model is itself a challenge.
Previous works circumvent this problem by defining human-likeness as \textit{"behavior that is indistinguishable from that of a human"} \cite{10.1093/mind/LIX.236.433, Turnwald2019, McIlroy-Young2020, Milani2023, He2024}. This definition enables two broad measurement categories: 
\begin{itemize}
    \item \textit{Quantitative tests}, which assess statistical deviations from observed human behavior.
    \item \textit{Qualitative tests}, often inspired by the Turing test \cite{10.1093/mind/LIX.236.433}, in which human observers evaluate behavioral data without knowing whether it was generated by a human or an artificial agent. These evaluations provide qualitative metrics on specific behavioral traits.
\end{itemize}

One such quantitative test fit for tabletop games was proposed in the paper introducing the Maia Chess engine \cite{McIlroy-Young2020} to assess the human-likeness of their model. It compares a set of observed decision-state pairs taken from the Lichess website's database with the model's decision in that same state.



A notable omission in their results is an ideal performance ceiling. Even among players of similar skill, individual decisions may diverge from the most common choice. Expecting perfect move-matching accuracy is unrealistic, but this ceiling can be estimated by measuring the dataset’s self-consistency. Additionally, while it quantifies behavioral similarity, it is not a standalone measure of human-likeness. Classification models trained on human data may perform well under this framework, but arguably lack generalization, raising questions about whether they truly exhibit human-like behavior.


Regarding qualitative tests, while the Turing test has been criticized as a measure of general intelligence \cite{French1990}, it remains effective when assessing targeted behavioral characteristics. For instance, Turnwald et al. \cite{Turnwald2019} asked human participants to rate their sense of comfort and cooperation when interacting with an agent, using human responses as a baseline for comparison with artificial agents. Rather than relying on a binary "human" vs. "non-human" classification, we aim to have participants rate specific agent qualities against a human baseline. For tabletop games, these qualities could include playstyle attributes such as aggressiveness, tactical depth, or the use of traps, alongside a general assessment of human-likeness.

Both qualitative and quantitative measurements are applicable to GGP, but their effectiveness is influenced by the diversity of games. Qualitative tests require observers familiar with human playstyles, while the accuracy of quantitative tests depends on the availability, quantity, and quality of human-generated data.

The challenge of obtaining sufficient human data can be mitigated using the Ludii database \cite{Crist2024}, which, in version 1.3.14, contains over 1,400 games and records approximately 100,000 anonymized plays per month from the Ludii community since the end of 2020. 

Qualitative testing is limited by the expertise of players and observers. For less popular games, evaluations may primarily reflect novice perspectives, reducing their reliability. In contrast, widely adopted board games such as Chess, Go, and Shogi offer access to expert players, making qualitative assessments more reliable for determining an agent’s general human-likeness.

These limitations should not prevent an accurate assessment of the human-likeness of our model, since it is not trained using game-specific human-generated data. Since the same underlying model is used across all games, validating its human-likeness in popular games—where more data and expert evaluations are available—supports the broader claim that it mimics human behavior consistently across all playable games.

\section{Conclusion and Future work}
This work introduces our vision for a human-like agent GGP model, based on results from cognitive psychology and past experiments in less specialized fields. We also highlight key challenges to the design of a robust, versatile and human-like artificial agent. 

Notably, existing measures of human-likeness are often imprecise and prone to false positives. A thorough investigation is needed to refine these metrics, enhance their accuracy, and explore alternative evaluation methods. We identified opponent modeling as a critical component of human problem-solving in adversarial contexts, but existing approaches rely on domain knowledge that may not be available in the context of GGP. This prompts the development of generalizable opponent-modeling methods.

Our future work includes an implementation of CogniPlay for the game of Renju, to first evaluate its performance within a single game. This will be followed by an implementation within the Ludii game-playing system to assess its generalizability.

\section*{Acknowledgment}
This article is based on the work of COST Action CA22145 - GameTable \cite{piette2024gametable}, supported by COST (European Cooperation in Science and Technology). 

\bibliography{cogniplay.bib}{}
\bibliographystyle{plain}

\end{document}